\documentclass[nohyperref]{article}

\usepackage{microtype}
\usepackage{graphicx}
\usepackage{subfigure}
\usepackage{booktabs} 

\usepackage{hyperref}

\usepackage[accepted]{icml2023}

\usepackage{amsmath}
\usepackage{amssymb}
\usepackage{mathtools}
\usepackage{amsthm}

\usepackage{amsmath,amsfonts,bm}

\def\eqref#1{equation~\ref{#1}}

\def\1{\bm{1}}

\DeclareMathAlphabet{\mathsfit}{\encodingdefault}{\sfdefault}{m}{sl}
\SetMathAlphabet{\mathsfit}{bold}{\encodingdefault}{\sfdefault}{bx}{n}

\usepackage{float}
\usepackage{hyperref}
\usepackage{url}
\usepackage{color}
\usepackage{algorithm}
\usepackage{algorithmic}
\usepackage{smartdiagram}

\usepackage[capitalize,noabbrev]{cleveref}

\theoremstyle{plain}
\newtheorem{theorem}{Theorem}[section]

\newtheorem{corollary}[theorem]{Corollary}
\theoremstyle{definition}

\newtheorem{assumption}[theorem]{Assumption}
\theoremstyle{remark}
\newtheorem{remark}[theorem]{Remark}

\usepackage[textsize=tiny]{todonotes}

\icmltitlerunning{Why the pseudo label based semi-supervised learning algorithm  is effective?}

\begin{document}

\twocolumn[
\icmltitle{Why the pseudo label based semi-supervised learning algorithm  is effective?}




\begin{icmlauthorlist}
\icmlauthor{Zeping Min}{yyy}
\icmlauthor{Qian Ge}{zzz}
\icmlauthor{Cheng Tai}{comp}
\end{icmlauthorlist}

\icmlaffiliation{yyy}{School of Mathematical Sciences,
Peking University, CHINA}
\icmlaffiliation{zzz}{Academy for Advanced Interdisciplinary Studies, Peking University, CHINA }
\icmlaffiliation{comp}{Moqi Technology
Zhongguancun International Innovation Building
Haidian District, Beijing, China}

\icmlcorrespondingauthor{Cheng Tai}{chengt@moqi.ai}

\icmlkeywords{Machine Learning, ICML}

\vskip 0.3in
]



\printAffiliationsAndNotice{} 

\begin{abstract}
Recently, pseudo label based semi-supervised learning has achieved great success in many fields. The core idea of the pseudo label based semi-supervised learning algorithm is to use models trained on labeled data to generate pseudo labels on  unlabeled data, and then train models to fit the previously generated pseudo labels. We give theoretical analysis of the effectiveness, convergence rate and sample complexity of  pseudo label based semi-supervised learning algorithm. 
We show that the pseudo label based semi-supervised learning algorithm is effective in the large sample regime where the number of unlabeled data goes to infinity, in which case the model which has the optimal population error upper bound.
More importantly, we give an explicit estimate of the rate of convergence achievable at each iteration. We also give the lower bound on sample complexity to achieve the target convergence rate. Experiments show that our estimates are fairly tight. Our analysis contributes to understanding the empirical successes of pseudo label based semi-supervised learning.

\end{abstract}

\section{Introduction}

Neural networks often require a large amount of labeled data to train. Labeled data is usually very time-consuming and labor-intensive to obtain. However, unlabeled data is often less expensive to obtain. Therefore, semi-supervised learning has become popular in the field of deep learning. The key to the success of semi-supervised learning is to effectively use unlabeled data to obtain better models. \citep{kingma2014semi}, \citep{laine2016temporal}, \citep{sohn2020fixmatch}, \citep{xie2020self}, \citep{shu2018dirt}, \citep{zhang2019bridging} and \citep{laine2016temporal} have put a lot of effort into using unlabeled data.

\subsection{Pseudo label based semi-supervised learning algorithm}
\label{section1.1}
In general, pretraining and generating pseudo labels are two main ways to use unlabeled data. Well-known pretrain models include \citep{devlin2018bert}, \citep{brown2020language}, \citep{baevski2020wav2vec} and \citep{liu2019roberta}. In this paper, we focus on pseudo label based semi-supervised learning algorithms \citep{grandvalet2004semi} and \citep{lee2013pseudo}.
The core idea of the pseudo label based semi-supervised learning algorithm is to use the model trained on the labeled data to generate pseudo labels on the unlabeled data, and then train a model to fit the previously generated pseudo labels. The sketch of pseudo label based semi-supervised learning algorithm is shown in Figure 1.
\begin{algorithm}
        \label{sketch}
	\renewcommand{\algorithmicrequire}{\textbf{Input:}}
	\renewcommand{\algorithmicensure}{\textbf{Output:}}
	\caption{Pseudo label based semi-supervised learning algorithm}
	\label{alg1}
	\begin{algorithmic}[1]
		\STATE Input: proper initial model $f_0$, unlabeled data $\mathcal{T}$, $i=0$, iteration number $I$.
		\REPEAT
	
		\STATE Generate pseudo label on $\mathcal{T}$ using $f_i$.
            \STATE Train on pseudo labeled $\mathcal{T}$ and  get $f_{i+1}$.
            \STATE $i \leftarrow i + 1$
		\UNTIL  $i = I$ 
		\ENSURE  Model $f_I$
	\end{algorithmic}  
\end{algorithm}

\begin{figure}[!htb]
\label{sketch2}
\begin{center}
\includegraphics[scale=0.4]{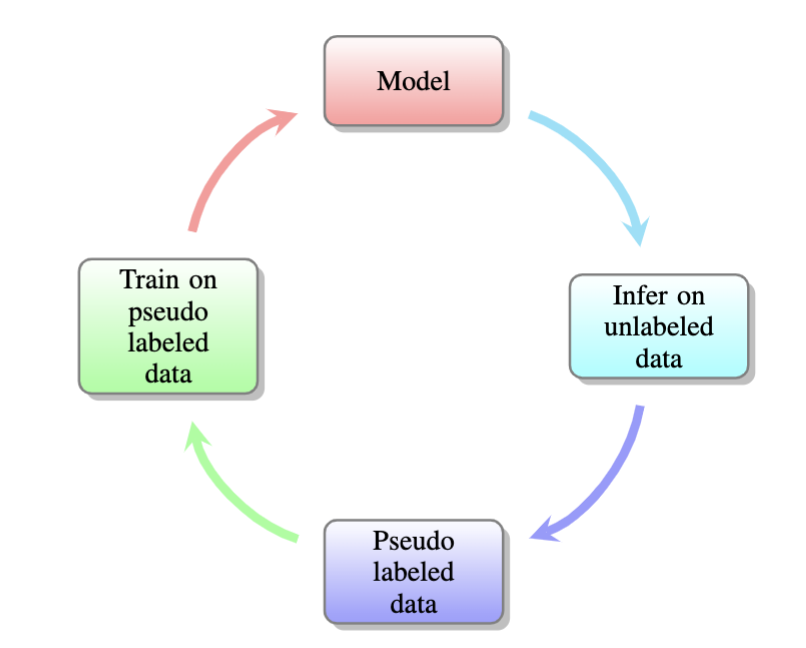}
\end{center}
\caption{Pseudo label based semi-supervised learning algorithm sketch}
\end{figure}

\subsection{Insight}\label{insight}
We provide a novel theoretical analysis of pseudo label based semi-supervised learning algorithm. Under a simple and realistic  assumption on the model, we show that when the amount of unlabeled data tends to infinity, the pseudo label based semi-supervised learning algorithm can obtain a model which has the same population error upper bound as supervised learning. More importantly, we give an explicit estimate of the rate of convergence achievable at each iteration. We also give the lower bound on sample complexity to achieve the target convergence rate.

Our assumption about the model is that if we train the model on  dataset part of which is  randomly labeled,  when the proportion of randomly labeled data is low, the model we get can have a lower empirical error on the correct labeled data, and a higher empirical error on the randomly labeled data. This is reasonable and verifiable especially when the model is under-parameterized. It is worth mentioning that even when the DNN models are over-parameterized, the DNN models still tend to fit correct data before mislabeled data.\citep{liu2020early} and \citep{arora2019fine}. We also show the intuition behind the assumption with a toy example. In this toy example, the dataset consists of $(x_1,y_1), (x_2,y_2), (x_3,y_3)$ (in color green), and a small part of  random mislabeled of them (in color orange). So when we train the DNN models on the dataset as in Figure 2, the model we get will ignore the mislabeled data.
\begin{figure}[htb]
\label{assumption}
\begin{center}
\includegraphics[width=0.5\textwidth]{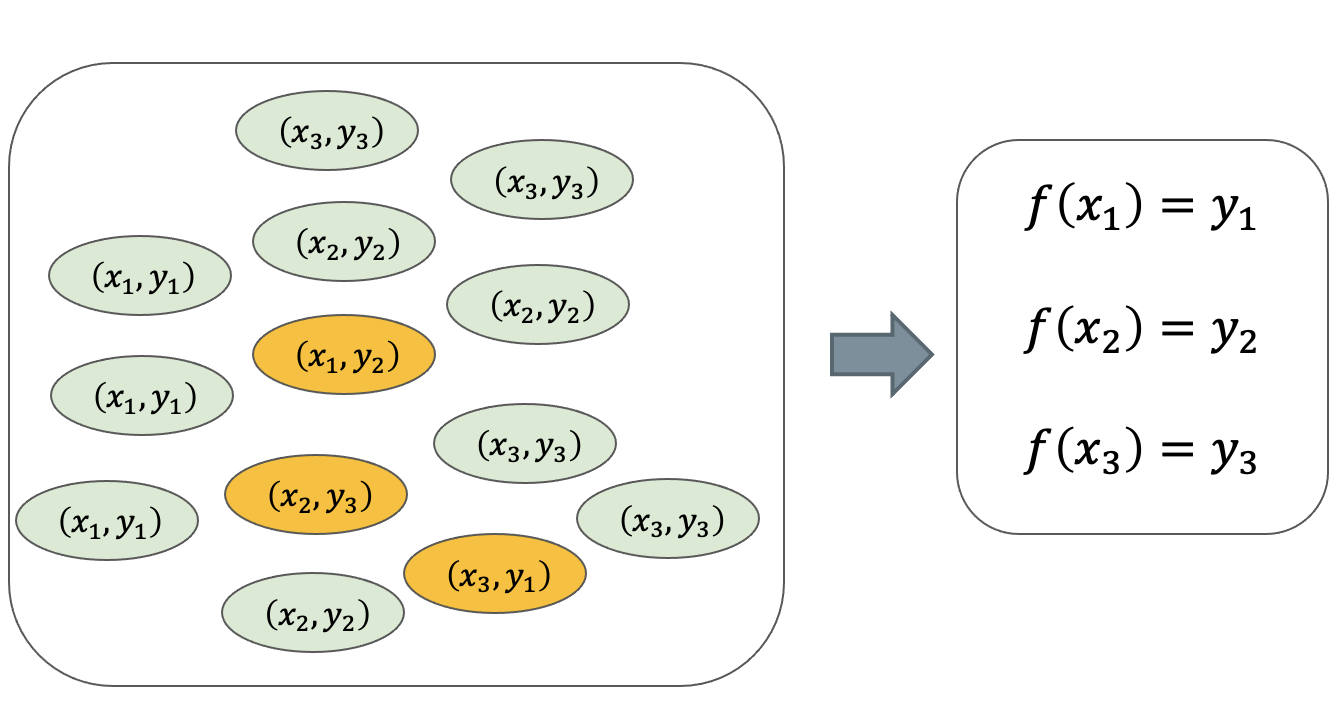}
\end{center}
\caption{A toy example for our assumption}
\end{figure}

The same intuition carries to more general cases. If we have a proper initial model and use it to generate pseudo labels on the unlabeled dataset, there will be a small part of the pseudo labels that are wrong-labeled. And if we use the generated pseudo label to train a new model, the model will try to fit the correct labels and  will be well-trained since the absolute quantity of the pseudo labels is large based on our assumption. What is more, if the model trained on the pseudo labels is good enough, we can reuse it to generate new pseudo labels on the unlabeled dataset. We hope that the pseudo labels generated by the previous model trained on the pseudo labels will have fewer wrong labels. So we can get a better model from them and the  iteration can continue. The process can be continued until no obvious improvement can be obtained. From this perspective, our analysis contributes to understanding the empirical successes
of pseudo label-based semi-supervised learning \citep{laine2016temporal}, \citep{tarvainen2017mean}, \citep{lee2013pseudo}, \citep{li2019speechtransformer}, \citep{graves2012sequence}, \citep{chiu2018state} and \citep{bengio2015scheduled}.

In summary, our contributions include:
\begin{itemize}
    \item We give a theoretical analysis of the pseudo label based algorithm and contribute to understanding the empirical successes of pseudo label based semi-supervised learning.
    \item We show that, if we have a proper initial model and when the amount of unlabeled data tends to infinity, the  algorithm can obtain the model which has the optimal population error upper bound. Here the optimal population error upper bound represents  the population error upper bound of the model obtained by supervised learning with all unlabeled data labeled.
    \item We give an explicit estimate of the rate of convergence achievable at each iteration. Besides, we also give the lower bound on sample complexity to achieve the target convergence rate. Experiments show that our estimates are fairly tight.
\end{itemize}

\section{Related work}
\subsection{Theory on pseudo label based semi-supervised learning}
In the early stage of machine learning, \citep{sain1996nature} proposes transductive SVM which tried to utilize the unlabeled data. Then \citep{derbeko2003error} estimates error bounds for transduction learning. Later, \citep{oymak2020statistical} shows  pseudo label based semi-supervised learning iterations improve model accuracy even though the model may be plagued by suboptimal fixed points. 
\citep{chen2020self} shows that, for a certain class of distributions, entropy minimization on unlabeled target data will reduce the interference of fake features. However, the analysis in \citep{oymak2020statistical} and \citep{chen2020self} mainly focus on linear models, and DNN models are not analyzed. For DNN models,  \citep{wei2020theoretical} shows that  pseudo label based semi-supervised learning method is beneficial to improve the performance of the DNN models, and gives a sample complexity. However, \citep{wei2020theoretical} did not show that the  pseudo label based semi-supervised algorithm can achieve the optimal population error, nor does it estimate the convergence rate of each iteration. Our work successfully addresses these issues.

\subsection{Population risk estimation method}
\label{RATT}
There are lots of methods to estimate the population risk of  DNN models. One of the most important methods is estimating the upper bound on the population error of DNN models by estimating the complexity of the hypothesis classes \citep{neyshabur2015norm}, \citep{neyshabur2017exploring}, \citep{ma2018priori} and \citep{weinan2019priori}. However, this method often is restricted to a specific model and it is hard to  use it to create a unified analysis to illustrate the advantage of pseudo label based semi-supervised learning for DNN models. Recently, \citep{garg2021ratt} established a method to estimate the population risk of the DNN models via the model performance on randomly labeled data. In the method of \citep{garg2021ratt}, we will not need to estimate the complexity of the hypothesis classes. So it can help us create a unified analysis for DNN models. And  using this method is very convenient to show the benefit of pseudo label based semi-supervised learning since there often will be some mislabeled data in the pseudo labels.

\section{Preliminary}
\subsection{Notation}
To be clear, we first show the notation in our paper. We mainly focus on the $k$ classification problem. Using $\mathcal{S}$ represents the labeled data, $n$ represents the amount of dataset  $\mathcal{S}$, $\tilde{\mathcal{S}} $ represents the randomly labeled data and $m$ represents the amount of dataset in $\tilde{\mathcal{S}}$. Using $\mathcal{E}_{\mathcal{S}}$  represents 0-1 loss on $\mathcal{S}$, $\mathcal{E}_{\tilde{\mathcal{S}}}$ represents 0-1 loss on $\mathcal{\tilde{S}}$, $\mathcal{E}_{\mathcal{D}}$ represents popluation 0-1 loss.
\subsection{Population risk upper bound estimation}
As described in Section \ref{RATT}, estimating population risk upper bound based on randomly labeled data is convenient to show the benefit of pseudo label based semi-supervised learning since there often will be some mislabeled data in the pseudo labels. What's more, it can help us to make a unified analysis for DNN models. Now we describe the theorem. This is obviously crucial for our following analysis.
\begin{assumption}
\label{assumption1}
Let $\hat{f}$ be a model obtained by training with an algorithm $\mathcal{A}$ on a mixture of clean data $S$ and randomly labeled data $\tilde{\mathcal{S}}$. Then with probability $1-\delta$ over the (uniform but without the correct label) mislabeled data $\tilde{\mathcal{S}}_M$, we assume that the following condition holds:
\begin{equation}
    \mathcal{E}_{\tilde{\mathcal{S}}_M}(\widehat{f}) \le \mathcal{E}_{\mathcal{D}^{\prime}}(\widehat{f})+c \sqrt{\frac{\log (1 / \delta)}{2 m}}
\end{equation}
for a fixed constant $c>0$. Where the $\mathcal{E}_{\mathcal{D}^{\prime}}(\widehat{f})$ represents the population loss of $\widehat{f}$ on (uniform but without the correct label) mislabeled data. \citep{garg2021ratt}
\end{assumption}
\begin{theorem}
\label{theorem1}
Under the Assumption \ref{assumption1}, then for any $\delta>0$, with probability at least $1-\delta$, we have

\begin{equation}
\mathcal{E}_{\mathcal{D}}(\widehat{f}) \le \mathcal{E}_{\mathcal{S}}(\widehat{f})+(k-1) \left(1-\frac{k}{k-1} \mathcal{E}_{\tilde{\mathcal{S}}}(\widehat{f})\right) \\
+c \sqrt{\frac{\log \left(\frac{4}{\delta}\right)}{2 m}}
\end{equation}
\\
for some constant $c$ satisfy 
\begin{equation}
    c \le \left(2 k+\sqrt{k}+\frac{m}{n \sqrt{k}}\right)
\end{equation}
\\
Where $m$ represents the amount of dataset $\tilde{\mathcal{S}}$ and $n$ represents the amount of dataset $\mathcal{S}$. \citep{garg2021ratt}
\end{theorem}
\begin{remark}
The randomly labeled data is not equal to mislabeled data. For $k$ classification problem, the randomly labeled data means for any $x$, its label is uniformly randomly selected from $k$ labels $y_1, y_2, y_3, ..., y_k$. However, the mislabeled data means for any $x$, its label is uniformly randomly selected from all $k$ labels except its ground truth label. For example, for $x_1$, and we suppose its ground truth label is $y_1$, then mislabeled data of $x_1$ is uniformly randomly selected from $k-1$ labels $y_2, y_3, ..., y_k$.
\end{remark}

\begin{remark}
    The Assumption \ref{assumption1}  holds in almost all scenarios. Since when we train DNN models, they always tend to overfit the training data. In practice, we often need to take steps to prevent overfitting.
\end{remark}

\subsection{General assumption}
\label{section3.3}
As we describe in Section \ref{insight}, our assumption about the model is that if we train the model on the dataset, which parts of it are randomly labeled. When the proportion of randomly labeled data is low, we can obtain models with low empirical error on correctly labeled data and high empirical error on randomly labeled data. Here we give a mathematical formula that describes this assumption in Assumption \ref{assumption2}.

\begin{assumption}
\label{assumption2}
$\exists \varepsilon>0, \tilde{\delta}>0, b \ge 0 $ if the training dataset $\mathcal{S} \cup \widetilde{\mathcal{S}}$  satisfy 
\begin{equation}
    \frac{m}{n} \le \widetilde{\delta}<1 
\end{equation}

we can get $\hat{f}$ that satisfy 
\begin{equation}
    \mathcal{E}_{\mathcal{S}}(\hat{f}) \le \varepsilon 
\end{equation}

\begin{equation}
    \mathcal{E}_{\tilde{\mathcal{S}}}(\hat{f}) \ge 1-\frac{1+b\varepsilon}{k}
\end{equation}

\end{assumption}
In the following section, we will go further under the condition of Assumption \ref{assumption2}. Since the $\varepsilon$ and $\tilde{\delta}$ change as the architecture of the model and training data change, exploring how the $\varepsilon$ and $\tilde{\delta}$ change as the architecture of the model and training data change is still an important work, and we think we will do it in the future.  And we will discuss under the fixed $\varepsilon$ and $\tilde{\delta}$ in this paper. 

In Section \ref{section4}, we discuss the  population risk under the condition that we have $N$ labeled data as the normal training setting.
In Section \ref{section5}, we show that when the amount of unlabeled data tends to infinity, the  algorithm can obtain the model which has the optimal population error upper bound.
In Section \ref{section6}, We give an estimation of the convergence rate and the sample complexity.

\section{Supervised learning}
\label{section4}
Firstly, according to Assumption \ref{assumption2}, we have 
\begin{equation}
    \frac{m}{n} \le \widetilde{\delta}<1 
\end{equation}
and we can give a more relaxed upper bound in Theorem \ref{theorem1} to simplify our analysis and notation later.

\begin{theorem}
\label{theorem2}
Under the Assumption \ref{assumption1}, then  for any $\delta>0$, with probability at least $1-\delta$, we have
\begin{equation}
\label{eq8}
\begin{aligned}
\mathcal{E}_{\mathcal{D}}(\widehat{f}) \le \mathcal{E}_{\mathcal{S}}(\widehat{f})+(k-1) &\left(1-\frac{k}{k-1} \mathcal{E}_{\tilde{\mathcal{S}}}(\widehat{f})\right) \\
+& ak \sqrt{\frac{\log \left(\frac{4}{\delta}\right)}{ m}}
\end{aligned}
\end{equation} 
with constant 
\begin{equation}
    ak \ge \frac{2k+\sqrt{k}+\frac{\widetilde{\delta}}{\sqrt{k}}}{\sqrt{2}} 
\end{equation}
and generally,  we can have $a=4$. 
Where $m$ represents the amount of dataset $\tilde{\mathcal{S}}$ and $n$ represents the amount of dataset $\mathcal{S}$ \citep{garg2021ratt}.

Specifically, if  the  $\widetilde{\delta}$ in Assumption \ref{assumption1} satisify 
\begin{equation}
\label{eq10}
    {2k+\sqrt{k}+\frac{\widetilde{\delta}}{\sqrt{k}}} < 2\sqrt{2}k
\end{equation}
We can have a tighter constant $a=2$ in equation \ref{eq8}. 

\end{theorem}
Now, we want to estimate the population risk upper bound of the models normal training on $N$ labeled data by Theorem \ref{theorem2}. On the one hand, we observe that the second term in Theorem \ref{theorem2} is related to  performance on random labeled data $\tilde{\mathcal{S}}$. And the third term $ak \sqrt{\frac{\log \left(\frac{4}{\delta}\right)}{ m}}$ decreases as $m$ increases. On the other hand, in Assumption \ref{assumption2}, the $\varepsilon$ show the model performance on both correct labeled data $\mathcal{S}$ and randomly labeled data $\tilde{\mathcal{S}}$. The $\widetilde{\delta}$ limit the upper bound of the amount of randomly labeled data $\tilde{\mathcal{S}}$.

Thus, we can randomly label a small part of $N$ labeled data and we can use the Theorem \ref{theorem2} to estimate the population risk and the randomly labeled small part won't affect much compared with training purely on $N$ labeled data. 
To satisfy Assumption \ref{assumption2}, we have 
\begin{equation}
\left\{
\begin{aligned}
&\frac{m}{n}=\tilde{\delta} \\
&m+n=N
\end{aligned}
\right.
\end{equation}
So we have
\begin{equation}
\left\{
\begin{aligned}
&m=\frac{\tilde{\delta}}{1+\tilde{\delta}} N \\
&n=\frac{1}{1+\tilde{\delta}} N.
\end{aligned}
\right.
\end{equation}

According to Assumption \ref{assumption2} and Theorem \ref{theorem2}, we have $f^{*}$ which satisfy 
\begin{equation}
\begin{aligned}
\mathcal{E}_{D}({f}^{*}) \le (1+b) \varepsilon+ak \sqrt{\log \left(\frac{4}{\delta}\right)} \frac{1}{\sqrt{\frac{\tilde{\delta}}{1+\tilde{\delta}} N}}
\end{aligned}
\end{equation}

We denote the population risk upper bound of $f^*$ as 
\begin{equation}
\label{eq12}
 \mathcal{E}_{D}^*:=(1+b) \varepsilon+ak \sqrt{\log \left(\frac{4}{\delta}\right)} \frac{1}{\sqrt{\frac{\tilde{\delta}}{1+\tilde{\delta}} N}}
\end{equation}
 which reflects the population risk upper bound under the normal setting in which we have  $N$ labeled training dataset.
 
 \begin{remark}
 We observe that the $\mathcal{E}_{D}^*$ is nearly optimal since it has the error rate order of $O(\frac{1}{\sqrt{N}})$ which is equal to Monte Carlo estimation error rate order. This also implies the reasonableness of our assumption.
 \end{remark}

\section{Effectiveness analysis}
\label{section5}
In this section, we first discuss the population risk under the condition that we have $N$ unlabeled data and a proper initial model as the pseudo label based algorithm setting. Then we compare the population risk in Section \ref{section4} and Section \ref{section5} as the amount of (unlabeled) data tends to infinity. We show that when the amount of unlabeled data tends to infinity, the pseudo label based semi-supervised learning algorithm can obtain a model with the same population error upper bound as the model obtained by supervised training in the condition of the amount of labeled data tends to infinity. 

\subsection{Modified pseudo label based algorithm}
\label{section5.1}
In this section we consider the case where we have $N$ unlabeled data and an initial $f_0$ as the  pseudo label based semi-supervised learning
algorithm setting as described in Section \ref{section1.1}. 

We need to generate the pseudo labels by the $f_0$ and then train the model by pseudo labels. We denote the $\mathcal{E}_{D}(f_0)$ as $\gamma_0$. Obviously, there are about $(1-\gamma_0)N$ correct labels and $\gamma_0N$ wrong labels in the generated pseudo labels. However, we can not view wrong labels in the generated pseudo labels as random label. A shred of direct evidence is there are no correct labels in the  wrong labels in the generated pseudo labels, but there are around $\frac{1}{k}$ correct labels in the random labels. Since both Assumption \ref{assumption2} and Theorem \ref{theorem2} connect to the model performance in randomly labeled data. A correct way is that we can select a small part of generated pseudo labels and then randomly label them. Then we can use Assumption \ref{assumption2} and Theorem \ref{theorem2} to estimate the population risk.
To be more clear, the modified pseudo label based algorithm, which is used to analyze is shown in Algorithm 2. Step 5 in Algorithm 2, which is mainly modified compared with Algorithm 1,  has little effect on pseudo label based algorithm in Algorithm 1 because in practice the $m<<N$ and we will show how to determine $m$ below. But the Algorithm 2 format can help us analyze using the population error tools mentioned above.

\begin{algorithm}[htb]
    \label{sketch3}
	\renewcommand{\algorithmicrequire}{\textbf{Input:}}
	\renewcommand{\algorithmicensure}{\textbf{Output:}}
	\caption{Modified pseudo label based  algorithm}
	\label{alg2}
	\begin{algorithmic}[1]
		\STATE Input: initial model $f_0$, $N$ unlabeled data $\mathcal{T}$, test data $\mathcal{T}_{test}$, $\varepsilon,\widetilde{\delta}$ that satisfy Assumption 2, iteration number $I$, $i=0$.
		\REPEAT
		
		\STATE Estimate the $f_i$ population risk $\gamma_i$  on $\mathcal{T}_{test}$
		\STATE Generate pseudo label on $\mathcal{T}$ using $f_i$.
		\STATE Random select proper $m (m << N)$ part of  data from pseudo labeled $\mathcal{T}$ and random labeled them 
        \STATE Update the pseudo labeled $\mathcal{T}$.
            \STATE Train on pseudo label $\mathcal{T}$. 
            \STATE $i \leftarrow i + 1$
		\UNTIL  $i = I$ 
		\ENSURE  Model $f_I$
	\end{algorithmic}  
\end{algorithm}

\subsection{Effectiveness of the pseudo label based
semi-supervised learning algorithm}
To satisfy Assumption \ref{assumption2}, the amount of data selected to random label $m$ in Algorithm 2 has the following restriction.

\begin{equation}
\begin{aligned}
\frac{m+\gamma_0(N-m)}{(1-\gamma_0)(N-m)} \le \widetilde{\delta} 
\end{aligned}
\end{equation}
So, we have

\begin{equation}
\label{eq14}
\begin{aligned}
&m \le \frac{\widetilde{\delta}(1-\gamma_0)-\gamma_0}{(1+\widetilde{\delta})(1-\gamma_0)} N
\end{aligned}
\end{equation}

\begin{equation}
\label{eq15}
\begin{aligned}
&\gamma_0 \le \frac{\widetilde{\delta}}{1+\tilde{\delta}}
\end{aligned}
\end{equation}

The equation \ref{eq14}  shows that we can select  at most $\frac{\widetilde{\delta}(1-\gamma_0)-\gamma_0}{(1+\widetilde{\delta})(1-\gamma_0)} N$ data from $N$ generated pseudo labels then random labeled them when the $\gamma_0$ satisfy equation \ref{eq15}. According to Assumption \ref{assumption2} and Theorem \ref{theorem2}, we can get $f_1$ and with at least $(1-\delta)$ probability we have
\begin{equation}
\mathcal{E}_{D}({f}_1) \le (1+b) \varepsilon+a k \sqrt{\log \left(\frac{4}{\delta}\right)} \sqrt{\frac{(1+\widetilde{\delta})(1-\gamma_0)}{\tilde{\delta}(1-\gamma_0)-\gamma_0}} \frac{1}{\sqrt{N}}
\end{equation}

Compare $\mathcal{E}_{D}({f}_1)$ with $ \mathcal{E}_{D}^*:=(1+b) \varepsilon+ak \sqrt{\log \left(\frac{4}{\delta}\right)} \frac{1}{\sqrt{\frac{\tilde{\delta}}{1+\tilde{\delta}} N}}$ in equation \ref{eq12}, we can easily show that 
\begin{equation}
\lim_{N \to +\infty} \frac{\mathcal{E}_{D}(f_1)}{\mathcal{E}_{D}^*} \le 1.
\end{equation}

In summary, we have

\begin{theorem}
Under the condition of Assumption \ref{assumption2}, for fixed  $\varepsilon$, $\widetilde{\delta}$ and $\forall \delta \in (0,1)$, if we have $f_0$ with $\gamma_0:=\mathcal{E}_{D}(f_0) < \frac{\widetilde{\delta}}{1+\tilde{\delta}}$ and $N$ unlabeled data, then by Algorithm 2, with at least $(1-\delta)$ probability, we can get  $f_1$ 
that satisfies 
\begin{equation}
\lim_{N \to +\infty} \frac{\mathcal{E}_{D}(f_1)}{\mathcal{E}_{D}^*} = 1.
\end{equation}

\end{theorem}

This result implies that if we have a proper $f_0$ and the amount of input unlabeled data  tends to be infinite, pseudo label based semi-supervised  algorithm can obtain a model in which  the population error upper bound is optimal which means it is equal to the population error upper bound of the model trained in the condition of amount of  labeled data tends to infinite. Further, this can be achieved even in ONE iteration. This actually shows the power of  the pseudo label based semi-supervised  algorithm.

\section{Sample complexity and convergence rate}
\label{section6}
In this section, we give an explicit estimate of the rate of convergence achievable at each iteration. We also give the lower bound on sample complexity to achieve the target convergence rate.

As analysis in Section \ref{section5.1}, if we use an initial model with population risk $\gamma_0$, the population risk uppper bound of $f_1$ by Algorithm 2  with at least $(1-\delta)$ probability we have

\begin{equation}
\mathcal{E}_{D}({f}_1) \le (1+b) \varepsilon+a k \sqrt{\log \left(\frac{4}{\delta}\right)} \sqrt{\frac{(1+\widetilde{\delta})(1-\gamma_0)}{\tilde{\delta}(1-\gamma_0)-\gamma_0}} \frac{1}{\sqrt{N}}
\end{equation}

If the model $f_1$ trained on the pseudo labels is good enough, we can reuse it to generate new pseudo labels on the unlabeled dataset and then obtain the new model $f_2$. We hope that the pseudo labels generated by the $f_1$ model trained on the pseudo labels will have fewer wrong labels than $f_0$. So we can get a better model from them and the iteration can continue. Here, we are interested in if the population risk upper bound can approximate the ${\mathcal{E}_{D}^*}$ and how fast it is. So we should care if we can achieve

\begin{equation}
\label{eq20}
\frac{{\mathcal{E}_{D}(f_{i+1})}-\mathcal{E}_{D}^*}{{\mathcal{E}_{D}(f_i)}-\mathcal{E}_{D}^*} \triangleq p \le p^* 
\end{equation}

where $f_{i},f_{i+1}$ denote the output of $i$th and $i+1$th iteration in Algorithm 2, $p^*$ is the target convergence rate in $(0,1)$. We denote the population risk of $f_i$ as $\gamma_i$. Then according to the analysis in Section \ref{section5.1}, if 

\begin{equation}
\label{eq23}
\begin{aligned}
&\gamma_i \le \frac{\widetilde{\delta}}{1+\tilde{\delta}}
\end{aligned}
\end{equation}

we can get $f_{i+1}$ that with at least $(1-\delta)$ probability we have

\begin{equation}
\begin{split}
\label{eq24}
\mathcal{E}_{D}({f}_{i+1}) &\le (1+b) \varepsilon+a k \sqrt{\log \left(\frac{4}{\delta}\right)} \sqrt{\frac{(1+\widetilde{\delta})(1-\gamma_i)}{\tilde{\delta}(1-\gamma_i)-\gamma_i}} \\ & \cdot\frac{1}{\sqrt{N}} \triangleq upper(f_{i+1})
\end{split}
\end{equation}

The straightforward point is that we have 


\begin{equation}
    \label{eq25}
    p \triangleq \frac{{\mathcal{E}_{D}(f_{i+1})}-\mathcal{E}_{D}^*}{{\mathcal{E}_{D}(f_i)}-\mathcal{E}_{D}^*}  \le \frac{upper(f_{i+1})-\mathcal{E}_{D}^*}{{\mathcal{E}_{D}(f_i)}-\mathcal{E}_{D}^*} \triangleq \widetilde{p}
\end{equation}
So we can consider equation \ref{eq20} by considering  
\begin{equation}
    \label{eq26}
    \widetilde{p} \le p^*
\end{equation}
Without loss of generality, we can further assume the $\gamma_i=\mathcal{E}_D(f_i)$ satisfies 

    \begin{equation}
    \mathcal{E}_D^*+c_1 \le \gamma_i \le \mathcal{E}_D^*+c_2
    \end{equation}
    
where $c_1$ and $c_2$ are two positive constant and $c_1$ can be arbitrarily small.

Then solve the equation \ref{eq26} we can get 
\begin{equation}
N \ge \left(\frac{a k}{p^* c_{1}}\right)^{2} \left[\sqrt{\frac{\tilde{\delta}+1}{\tilde{\delta}-\frac{\mathcal{E}_D^*+c_{2}}{1-\mathcal{E}_D^*-c_{2}}}}-\sqrt{\frac{\tilde{\delta}+1}{\tilde{\delta}}}\right]^{2}\log \left(\frac{4}{\delta}\right)
\end{equation}

Hence, we have 
\begin{theorem}[\textbf{Sample Complexity Estimation}]
    \label{th4}
    Under the condition of Assumption \ref{assumption2}, for fixed  $\varepsilon$, $\widetilde{\delta}$ satisfy Assumption \ref{assumption2} and $\forall 
    \delta, c_1, c_2, p^* \in (0,1)$ and $c_1 \le c_2$,  we  define 
     $\mathcal{E}_D^*$ as equation \ref{eq12}. If the number of unlabeled data  $N$ satisfy  
    \begin{equation}
    \label{eq27}
    N \ge \left(\frac{a k}{p^* c_{1}}\right)^{2} \left[\sqrt{\frac{\tilde{\delta}+1}{\tilde{\delta}-\frac{\mathcal{E}_D^*+c_{2}}{1-\mathcal{E}_D^*-c_{2}}}}-\sqrt{\frac{\tilde{\delta}+1}{\tilde{\delta}}}\right]^{2}\log \left(\frac{4}{\delta}\right)
    \end{equation} 
    and 
    \begin{equation}
    \mathcal{E}_D^*+c_1\le  \mathcal{E}_D(f_i)\le \mathcal{E}_D^*+c_2
    \end{equation}
    \begin{equation}
     \mathcal{E}_D(f_i) \le \frac{\tilde{\delta}}{1+\tilde{\delta}}
    \end{equation}
    then with at least $(1-\delta)^{2}$ probability  we have 
    \begin{equation}
    p \triangleq \frac{{\mathcal{E}_{D}(f_{i+1})}-\mathcal{E}_{D}^*}{{\mathcal{E}_{D}(f_i)}-\mathcal{E}_{D}^*}  \le p^* 
    \end{equation}
where $f_{i},f_{i+1}$ denote the output of $i$th and $i+1$th iteration in Algorithm 2.
\end{theorem}
Since $c_1$ could be arbitrarily small, this result indeed shows that the population risk by pseudo label based algorithm can  approximate to the normally trained population risk ${\mathcal{E}_{D}^*}$ as the pseudo label based algorithm iteration progresses. 

And note that for a fix $N$ and $p^*$, the right-hand term of equation \ref{eq27} increases as the $c_1$ decreases and as $c_2$ increases. Hence  for $f_i$, the most loose condition of equation \ref{eq27} will reach at $c_1=c_2=\mathcal{E}_{D}(f_{i})-\mathcal{E}_{D}^*$. So we have,

\begin{corollary}
    \label{co1}
    Under the condition of Assumption \ref{assumption2}, for fixed  $\varepsilon$, $\widetilde{\delta}$ satisfy Assumption \ref{assumption2} and $\forall 
    \delta, p^* \in (0,1)$,  we  define 
     $\mathcal{E}_D^*$ as equation \ref{eq12}. If the number of unlabeled data  $N$ satisfy  
    \begin{equation}
    \begin{split}
    N \ge & \left(\frac{a k}{p^* (\mathcal{E}_D(f_i)-\mathcal{E}_{D}^*)}\right)^{2} \left[\sqrt{\frac{\tilde{\delta}+1}{\tilde{\delta}-\frac{\mathcal{E}_D(f_i)}{1-\mathcal{E}_D(f_i)}}}-\sqrt{\frac{\tilde{\delta}+1}{\tilde{\delta}}}\right]^{2}\\
    &\cdot\log \left(\frac{4}{\delta}\right)
    \end{split}
    \end{equation} 
    and 
    \begin{equation}
     \mathcal{E}_D(f_i) \le \frac{\tilde{\delta}}{1+\tilde{\delta}}
    \end{equation}
    then with at least $(1-\delta)^{2}$ probability  we have 
    \begin{equation}
    p \triangleq \frac{{\mathcal{E}_{D}(f_{i+1})}-\mathcal{E}_{D}^*}{{\mathcal{E}_{D}(f_i)}-\mathcal{E}_{D}^*}  \le p^* 
    \end{equation}
where $f_{i},f_{i+1}$ denote the output of $i$th and $i+1$th iteration in Algorithm 2.
\end{corollary}

More importantly, we can estimate the convergence rate by equation \ref{eq24} and equation \ref{eq25} when we have a fixed number of samples $N$. 

\begin{theorem}[\textbf{Convergence Rate Estimation}]
    \label{th5}
    Under the condition of Assumption \ref{assumption2}, for fixed  $\varepsilon$, $\widetilde{\delta}$ satisfy Assumption \ref{assumption2} and $\forall 
    \delta \in (0,1)$,  we  define 
     $\mathcal{E}_D^*$ as equation \ref{eq12}. If we have   $N$ samples, and 
    \begin{equation}
     \mathcal{E}_D(f_i) \le \frac{\tilde{\delta}}{1+\tilde{\delta}}
    \end{equation}
    then with at least $O(1-\delta)^{2}$ probability  we have 
    \begin{equation}
    \begin{split}
    \label{eq37}
    p \le &\frac{(a k)}{ (\mathcal{E}_D(f_i)-\mathcal{E}_{D}^*) \sqrt{N}} 
    \left[\sqrt{\frac{\tilde{\delta}+1}{\tilde{\delta}-\frac{\mathcal{E}_D(f_i)}{1-\mathcal{E}_D(f_i)}}}-\sqrt{\frac{\tilde{\delta}+1}{\tilde{\delta}}}\right]\\
    &\cdot\sqrt{\log \left(\frac{4}{\delta}\right)}
    \end{split}
    \end{equation}

    Here $
    p \triangleq \frac{{\mathcal{E}_{D}(f_{i+1})}-\mathcal{E}_{D}^*}{{\mathcal{E}_{D}(f_i)}-\mathcal{E}_{D}^*}$ as above and  $f_{i},f_{i+1}$ denote the output of $i$th and $i+1$th iteration in Algorithm 2.
\end{theorem}

\section{Experiment}
\subsection{Dataset}
We conduct experiments on the CIFAR-10 dataset and FashionMnist dataset respectively. Both the CIFAR-10 dataset and the FashionMnist dataset are  widely used datasets in machine learning research. For CIFAR-10 dataset, it contains 60,000 32x32 color images from 10 different classes. The training split has 50,000 32x32 color images and the test split has 10,000 32x32 color images. For FashionMnist dataset, it covers a total of 70,000 28x28 grayscale images from 10 categories, of which 60,000 are used as training sets and 10,000 are used as test sets.

 In order to ensure the sample volume requirements, we expanded the sample size of the training set from 50,000 to 200,000 (4 times) by random cropping, etc for the CIFAR-10 dataset and expanded the sample size of the training set from 60,000 to 180,000 (3 times) by random cropping, etc for FashionMnist dataset. Besides, we construct a binary classification problem instead of an official 10 classification problem by combining five classes in the dataset into one class for both experiments on two datasets.
\subsection{Model}
\label{sec7.2}
\begin{figure*}[htb]
\begin{center}
\includegraphics[width=0.8\textwidth]{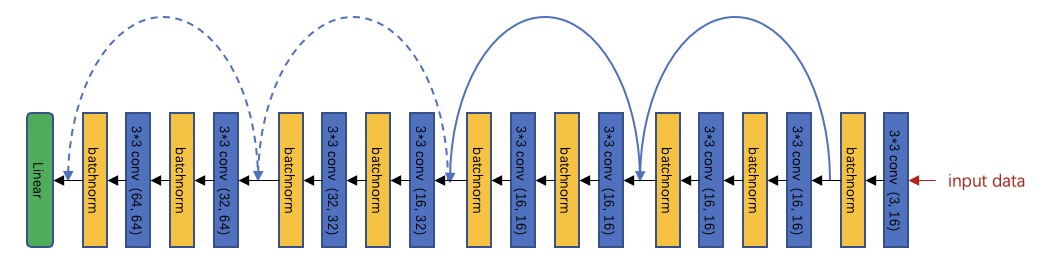}
\end{center}
\caption{Model architecture}
\label{resnet}
\end{figure*}

We use a 10 layers resnet \cite{he2016deep} with a 3x3 kernel size for both experiments on two datasets. Model structure and hyperparameters are shown in Figure \ref{resnet}. For optimization, we use the SGD optimizer with an initial 0.1 learning rate and a multistep learning rate scheduler.

\subsection{Experiment on CIFAR-10 dataset}
\label{sec7.3}

 We show $\tilde{\delta}=0.3, \varepsilon=0.0465, b=0.906$ satisfy Assumption \ref{assumption2} in Appendix \ref{sec7.3.1}. Now we test the convergence rate in Theorem \ref{th5}. In the experiment we have the number of classes $k=2$, hence equation \ref{eq10} is correct and we can select $a=2$. So according to the equation \ref{eq12}, we can calculate the $ \mathcal{E}_{D}^*:=(1+b) \varepsilon+ak \sqrt{\log \left(\frac{4}{\delta}\right)} \frac{1}{\sqrt{\frac{\tilde{\delta}}{1+\tilde{\delta}} N}}$ with different probability tolerance $1-\delta$. The results are shown in Table \ref{ta1}.

\begin{table}[htb]
\centering
\begin{tabular}{llll}
\hline
$\delta$ & 0.025  & 0.05 & 0.1  \\
\hline
$\mathcal{E}_{D}^*$   & 0.131 & 0.128 & 0.124 \\
\hline
\end{tabular}
\caption{The $ \mathcal{E}_{D}^*$ with different $\delta$}
\label{ta1}
\end{table}

For $f_i$ ($i=0$ here), we get the $f_i$ by training the model in subsection \ref{sec7.2} on 10000 images randomly sampled on raw  CIFAR-10 dataset and achieve a 0.183 test error rate. The error rate on dataset test  split via  epoch is shown in Appendix \ref{a2}.

\begin{figure}[ht]
\begin{center}
\includegraphics[width=0.5\textwidth,height=0.48\textwidth]{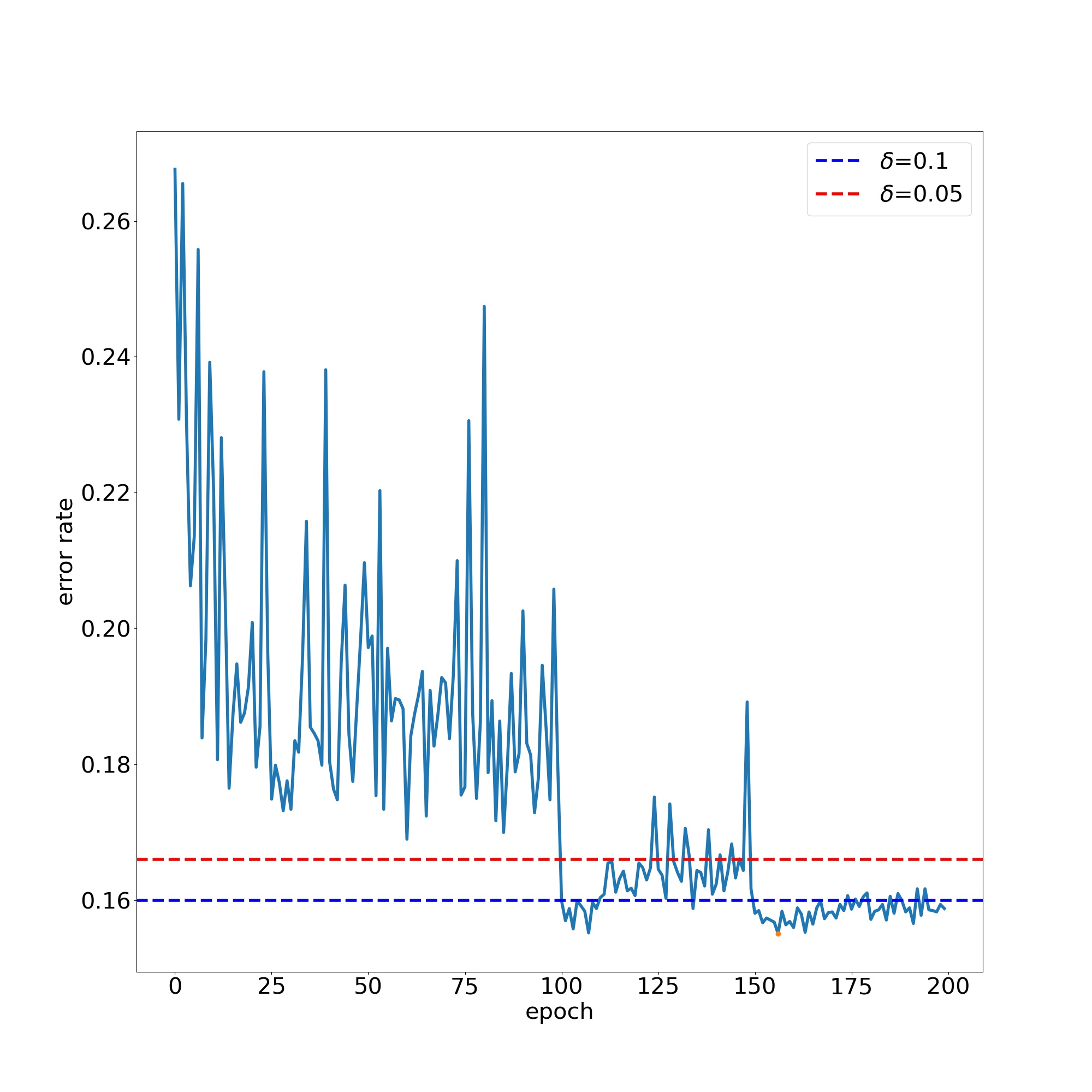}
\end{center}
\caption{The $\mathcal{E}_D(f_{i+1})$ upper bound with different  probability tolerance $(1-\delta)^2$: $\delta=0.05$ (red line) and $\delta=0.1$ (blue line). Hence the upper bound of convergence rate in Theorem \ref{th5} is fairly tight.}
\label{fig6}
\end{figure}

Finally, we can calculate the convergence rate according to equation \ref{eq37} with different  probability tolerance $(1-\delta)^2$. And then we can get the $\mathcal{E}_D(f_{i+1})$ upper bound with different  probability tolerance $(1-\delta)^2$ by $p \triangleq \frac{{\mathcal{E}_{D}(f_{i+1})}-\mathcal{E}_{D}^*}{{\mathcal{E}_{D}(f_i)}-\mathcal{E}_{D}^*}$. Experiment results are shown in Figure \ref{fig6}. It can be seen that the bounds we give are quite tight.

\subsection{Experiment on FashionMnist dataset}
We show $\tilde{\delta}=0.35, \varepsilon=0.024, b=0.77$ satisfy Assumption \ref{assumption2} in Appendix \ref{sec7.4}. Now we test the convergence rate in Theorem \ref{th5}. And in the experiment we have the number of classes $k=2$, hence equation \ref{eq10} is correct and we can select $a=2$. So according to the equation \ref{eq12}, we can calculate the $ \mathcal{E}_{D}^*:=(1+b) \varepsilon+ak \sqrt{\log \left(\frac{4}{\delta}\right)} \frac{1}{\sqrt{\frac{\tilde{\delta}}{1+\tilde{\delta}} N}}$ with different probability tolerance $1-\delta$. The resultas are shown in Table \ref{ta2}

\begin{table}[htb]
\centering
\begin{tabular}{llll}
\hline
$\delta$ & 0.025  & 0.05 & 0.1  \\
\hline
$\mathcal{E}_{D}^*$   & 0.084 & 0.081 & 0.078\\
\hline
\end{tabular}
\caption{The $ \mathcal{E}_{D}^*$ in different  $\delta$ selection }
\label{ta2}
\end{table}
For $f_i$ ($i=0$ here), we get the $f_i$ by training the model in subsection \ref{sec7.2} on 1200 images randomly sampled on the raw  training  set of FashionMnist dataset  and achieve a 0.095 test error rate. The error rate on dataset test  split via  epoch is shown in Appendix \ref{b2}.

\begin{figure}[htb]
\begin{center}
\includegraphics[width=0.5\textwidth,height=0.48\textwidth]{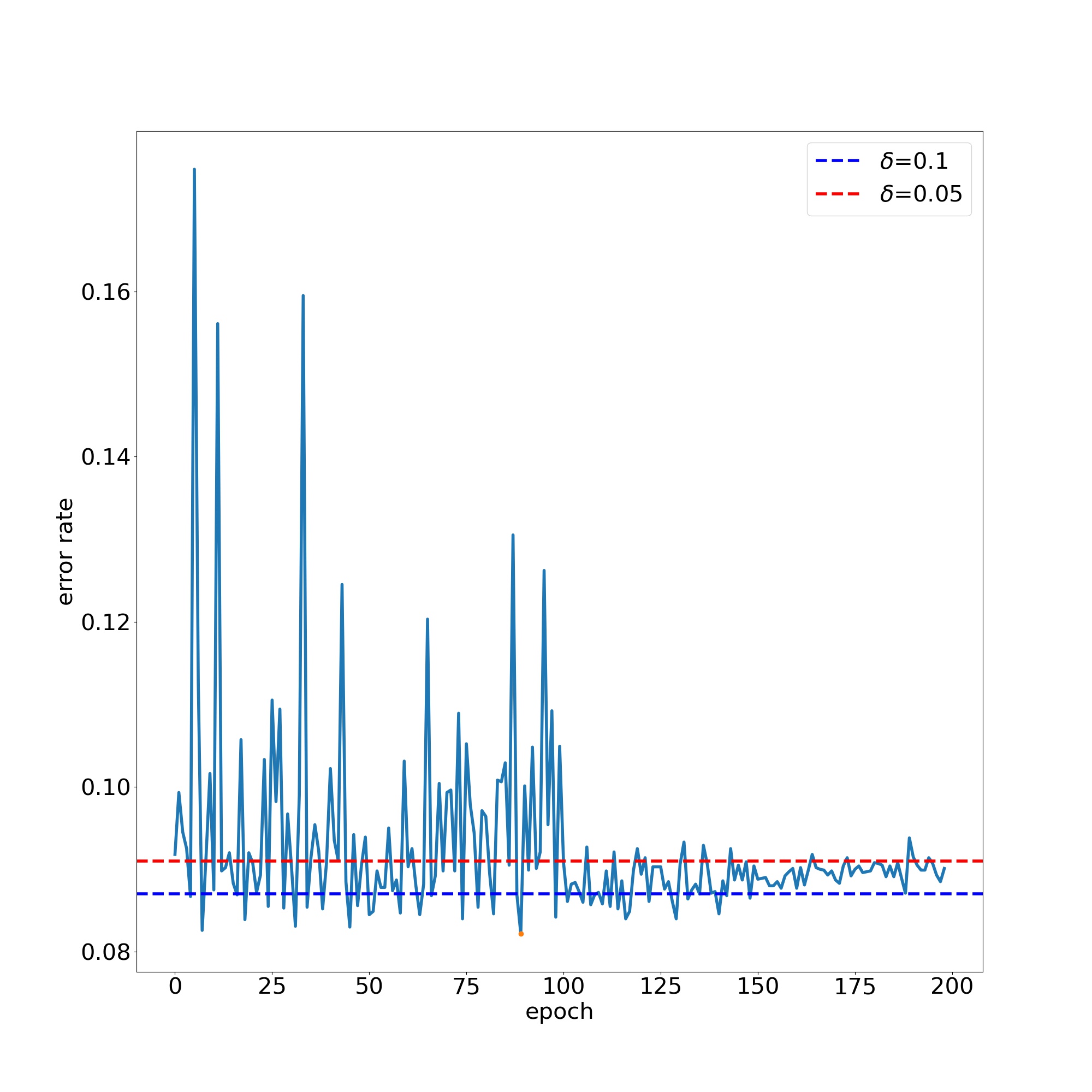}
\end{center}
\caption{The $\mathcal{E}_D(f_{i+1})$ upper bound with different  probability tolerance $(1-\delta)^2$: $\delta=0.1$ (blue line) and $\delta=0.05$ (red line). Hence the upper bound of convergence rate in Theorem \ref{th5} is fairly tight.}
\label{fig9}
\end{figure}

Finally, we can calculate the convergence rate according to equation \ref{eq37} with different  probability tolerance $(1-\delta)^2$. And then we can get the $\mathcal{E}_D(f_{i+1})$ upper bound with different  probability tolerance $(1-\delta)^2$ by $
    p \triangleq \frac{{\mathcal{E}_{D}(f_{i+1})}-\mathcal{E}_{D}^*}{{\mathcal{E}_{D}(f_i)}-\mathcal{E}_{D}^*}$. Experiment results are shown in Figure \ref{fig9}.

\section{Conclution and future work}

We conduct a theoretical analysis of pseudo label based algorithm. We analyzed its effectiveness and we give an explicit estimate of the rate of convergence and sample complexity. Our Assumption \ref{assumption2} is important to our analysis and we have explained the reasonableness of the assumption in Section \ref{insight}. But how the $\varepsilon$ and $\tilde{\delta}$ change as the architecture of the model and training data change is still mysterious to us, and we will explore it in the future. We hope that our analysis helps to understand the empirical success and reveal the potential of pseudo label based semi-supervised learning algorithm, and facilitate its application in wider scenarios.

\bibliography{icml2023}
\bibliographystyle{icml2023}
\newpage
\appendix
\onecolumn
\section{Experiment on CIFAR-10 dataset}
\subsection{Assumption test}
\label{sec7.3.1}
\begin{figure*}[!htbp]
\centering
\begin{minipage}[!htbp]{0.48\textwidth}
\centering
\includegraphics[width=\textwidth]{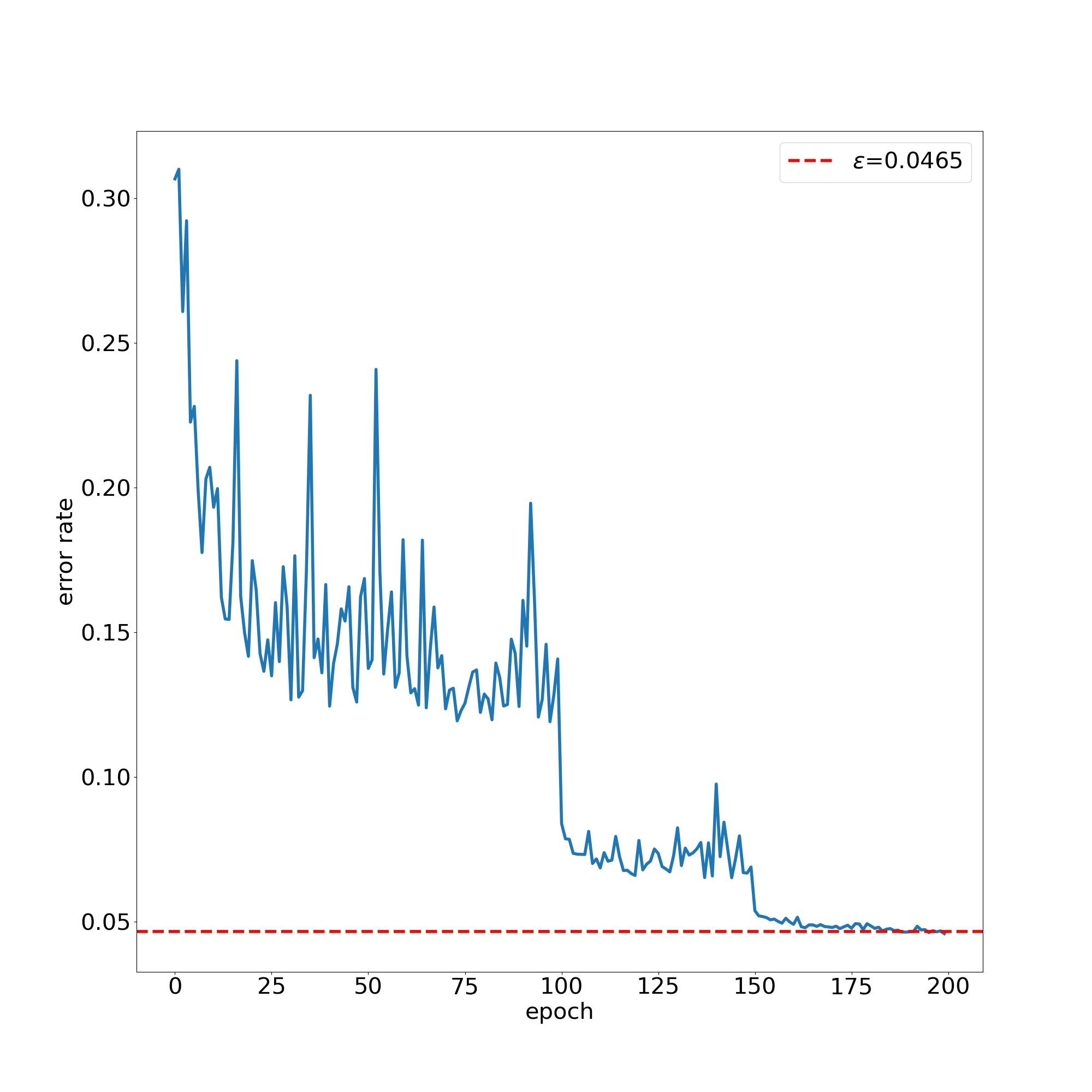}
\end{minipage}
\begin{minipage}[!htbp]{0.48\textwidth}
\centering
\includegraphics[width=\textwidth]{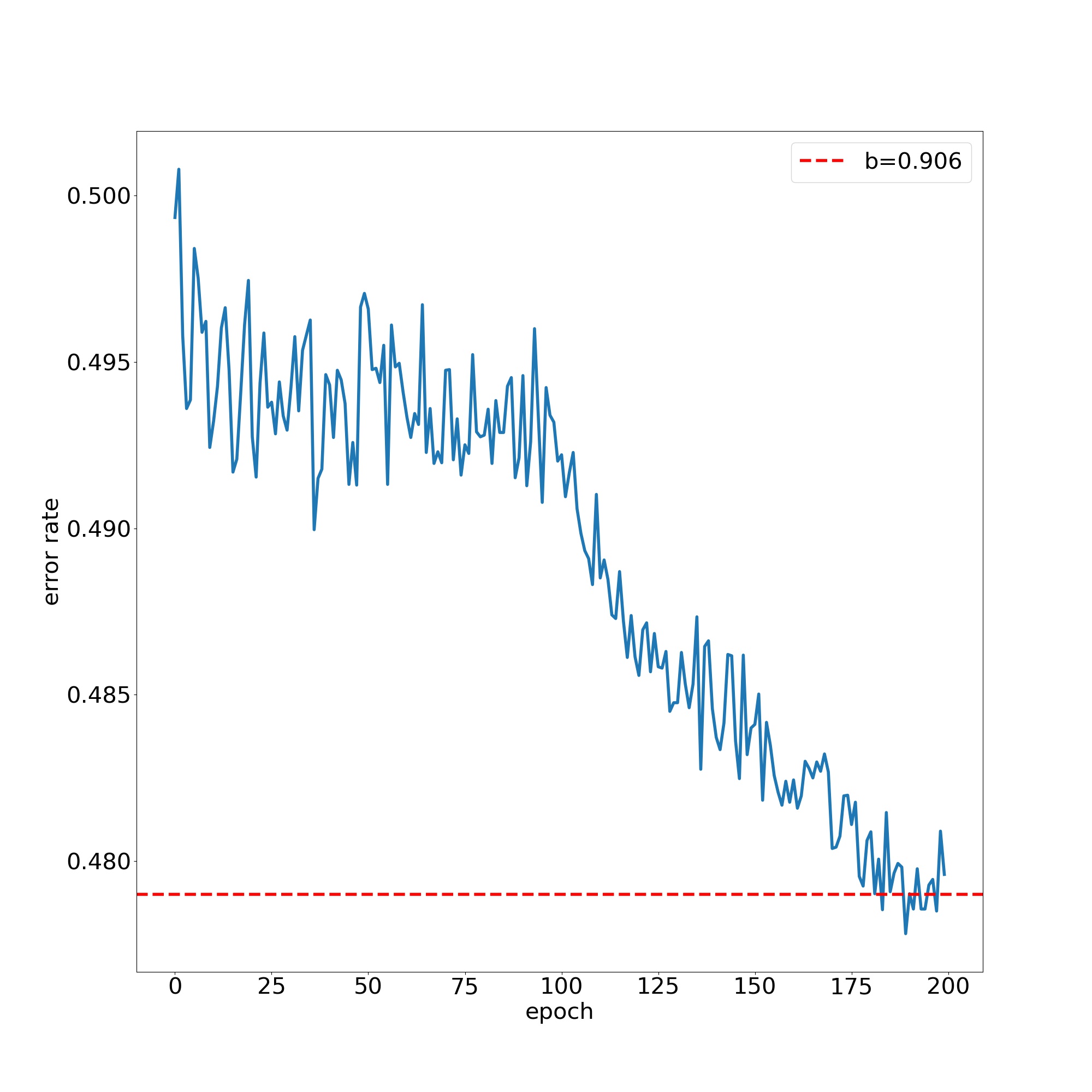}
\end{minipage}
\caption{Error rate on correct labeled data $\mathcal{S}$ (left) and random labeled data $\widetilde{\mathcal{S}}$ (right). We can see that the  $\varepsilon=0.0465, b=0.906$ is a good choice for Assumption \ref{assumption2}.}
\label{fig4}
\end{figure*}

We should determine the appropriate $\varepsilon, \tilde{\delta}, b $ in the Assumption \ref{assumption2}. We conduct this by selecting a proper $\tilde{\delta}$ and then determine the proper $\varepsilon$ and $b$ according to the experiment results under the $\tilde{\delta}$. Here, we select $\tilde{\delta}=0.3$ and hence $200,000 \times \tilde{\delta}=60,000$ are random labeled. As description in Assumption \ref{assumption2}, we train on $\mathcal{S} \cup \widetilde{\mathcal{S}}$ and the error rate on $\mathcal{S}$ and $\widetilde{\mathcal{S}}$ are shown in Figure \ref{fig4}. 
We first select the $\varepsilon=0.0465$ according to the error rate on $\mathcal{S}$ because  $\varepsilon=0.0465$ is relatively small compared with the model error rate and can lead to a tight upper bound. 
Then we select $b=0.906$ to make sure $1-\frac{1+b\varepsilon}{k}$ is a lower bound of     the error rate on $\tilde{\mathcal{S}}$. The red line in the right picture of Figure \ref{fig4} is $1-\frac{1+b\varepsilon}{k}$ with $b=0.906, \varepsilon=0.0465, k=2$ and we see that $b=0.906$ is a fairly good choice. In summary, we select $\tilde{\delta}=0.3, \varepsilon=0.0465, b=0.906$ which satisfy the Assumption \ref{assumption2}.

\begin{figure}[!htbp]
\begin{center}
\includegraphics[width=0.5\textwidth,height=0.48\textwidth]{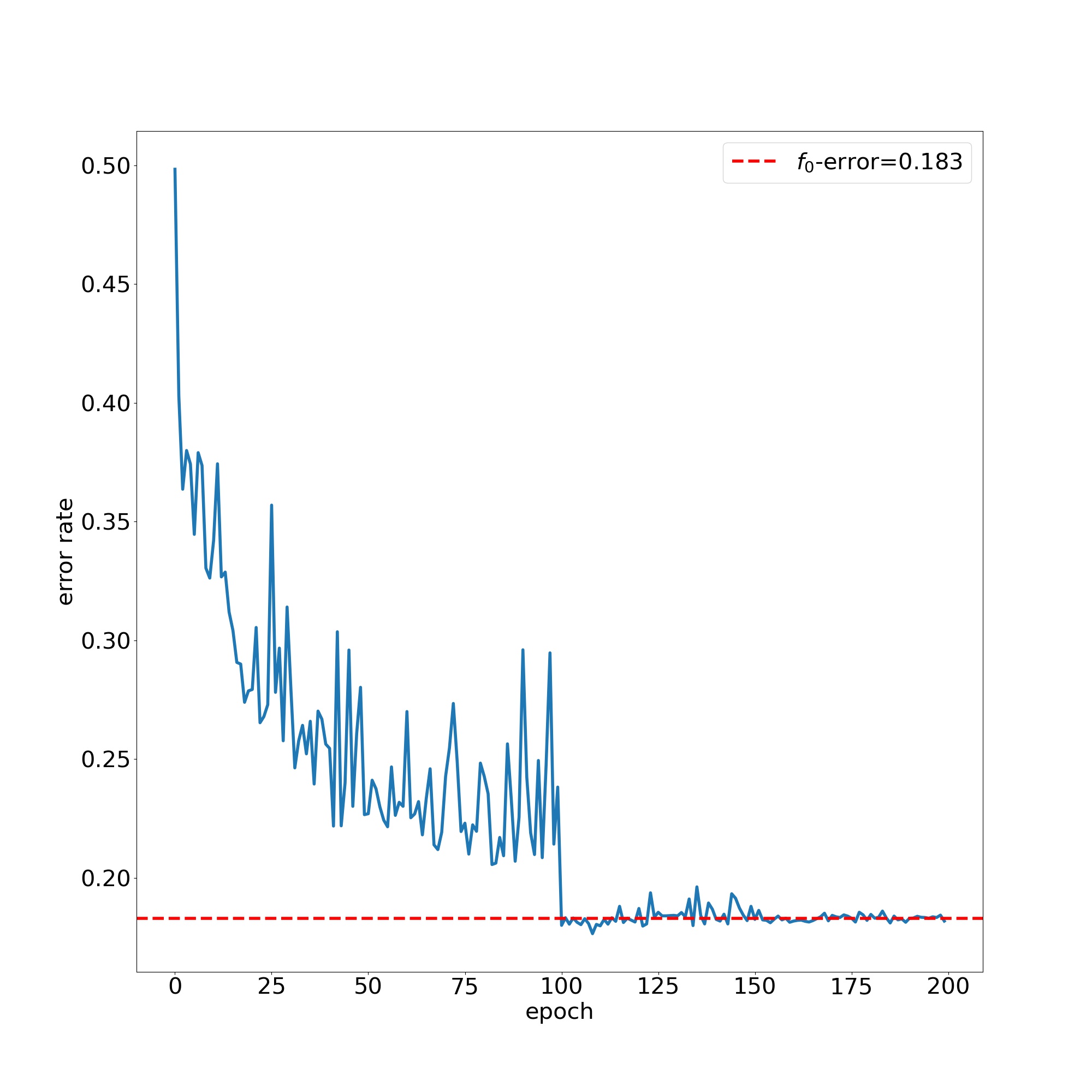}
\end{center}
\caption{The test error rate of $f_i$  via  epoch and we  can see that the $f_i$ can achieve a 0.183 test error rate.}
\label{fig5}
\end{figure}

\subsection{Model $f_i$}
\label{a2}
For $f_i$ ($i=0$ here), we get the $f_i$ by training the model in subsection \ref{sec7.2} on 10000 images randomly sampled on raw  CIFAR-10 dataset and achieve a 0.183 test error rate. The error rate on dataset test  split via  epoch is shown in Figure \ref{fig5}.

\section{Experiment on FashionMnist dataset}
\subsection{Assumption test}
\label{sec7.4}
\begin{figure*}[!htbp]
\centering
\begin{minipage}[t]{0.48\textwidth}
\centering
\includegraphics[width=\textwidth]{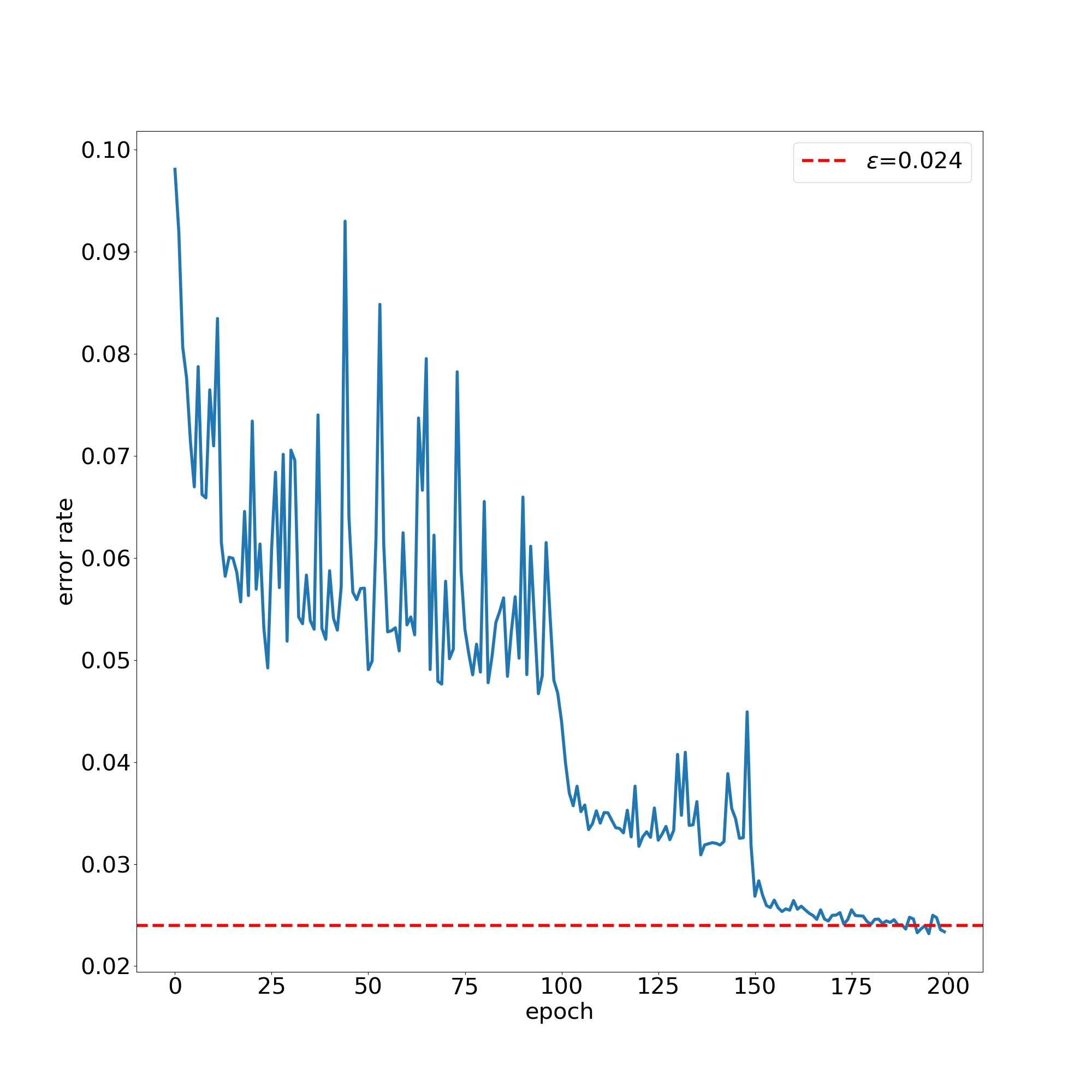}
\end{minipage}
\begin{minipage}[t]{0.48\textwidth}
\centering
\includegraphics[width=\textwidth]{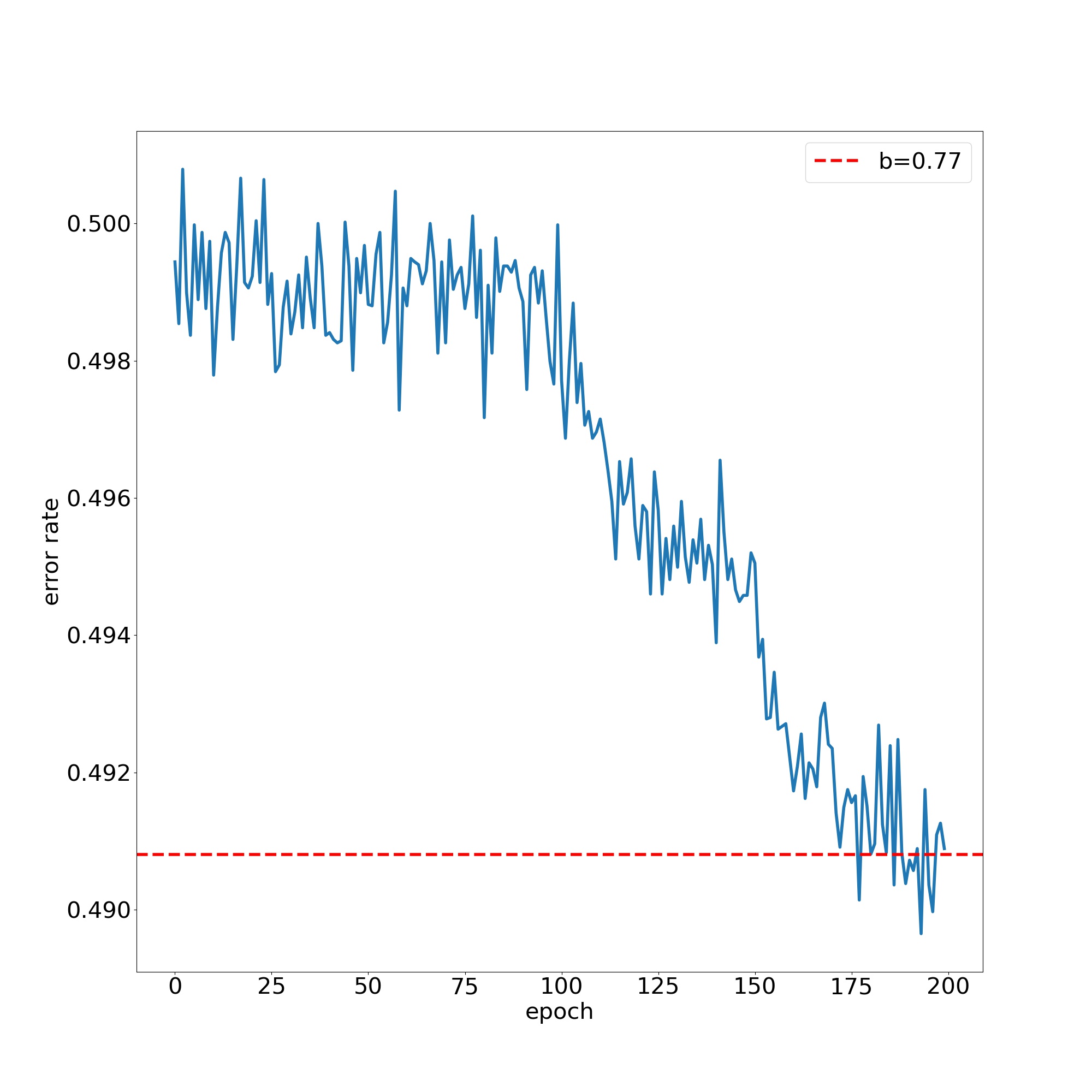}
\end{minipage}
\caption{Error rate on correct labeled data $\mathcal{S}$ (left) and random labeled data $\widetilde{\mathcal{S}}$ (right). We can see that the  $\varepsilon=0.024, b=0.77$ is a good choice for Assumption \ref{assumption2}.}
\label{fig7}
\end{figure*}

Similar to Section \ref{sec7.3}, firstly, we should determine the appropriate $\varepsilon, \tilde{\delta}, b$ in the Assumption \ref{assumption2}. We conduct this by selecting a proper $\tilde{\delta}$ and then determine the proper $\varepsilon$ and $b$ according to the experiment results under the $\tilde{\delta}$. Here, we select $\tilde{\delta}=0.35$ and hence $180,000 \times \tilde{\delta}=63,000$ are random labeled. As description in Assumption \ref{assumption2}, we train on $\mathcal{S} \cup \widetilde{\mathcal{S}}$ and the error rate on $\mathcal{S}$ and $\widetilde{\mathcal{S}}$ are shown in Figure \ref{fig7}. 
We first select the $\varepsilon=0.024$ according to the error rate on $\mathcal{S}$ because  $\varepsilon=0.024$ is relatively small compared with the model error rate on the $\mathcal{S}$ and can lead to a fairly tight upper bound. 
Then we select $b=0.77$ to make sure $1-\frac{1+b\varepsilon}{k}$ is a lower bound of     the error rate on $\tilde{\mathcal{S}}$. The red line in the right picture of Figure \ref{fig7} is $1-\frac{1+b\varepsilon}{k}$ with $b=0.77, \varepsilon=0.024, k=2$ and we see that $b=0.77$ is a fairly good choice. In summary, we select $\tilde{\delta}=0.35, \varepsilon=0.024, b=0.77$ which satisfy the Assumption \ref{assumption2}.

\subsection{Model $f_i$}
\label{b2}
\begin{figure}[!htbp]
\begin{center}
\includegraphics[width=0.5\textwidth,height=0.48\textwidth]{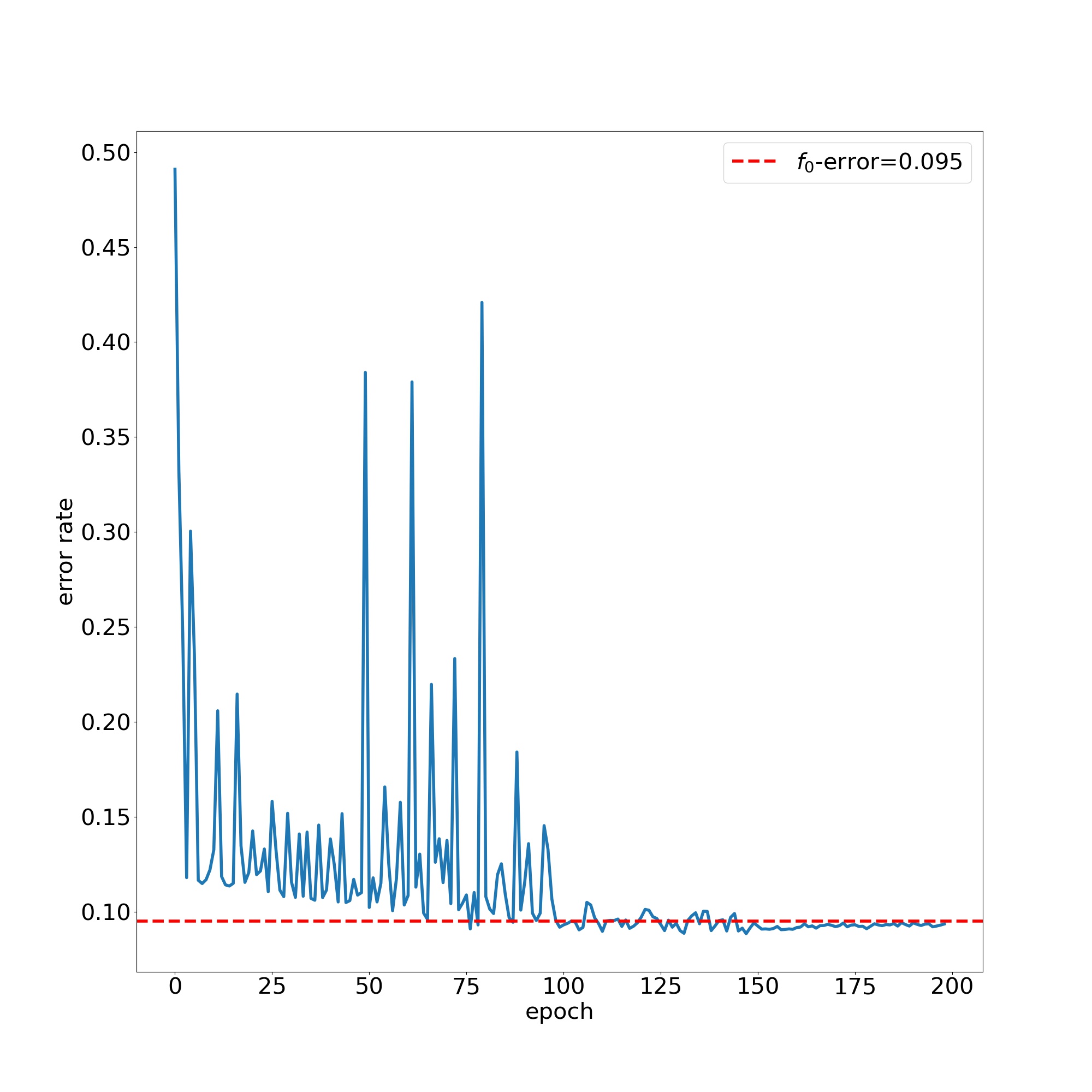}
\end{center}
\caption{The test error rate of $f_i$  via  epoch and we  can see that the $f_i$ can achieve a 0.095 test error rate.}
\label{fig8}
\end{figure}

For $f_i$ ($i=0$ here), we get the $f_i$ by training the model in subsection \ref{sec7.2} on 1200 images randomly sampled on the raw  training  set of FashionMnist dataset  and achieve a 0.095 test error rate. The error rate on dataset test  split via  epoch is shown in Figure \ref{fig8}. 

\end{document}